\documentclass[utf8]{frontiersSCNS} 

\usepackage{url,hyperref,lineno,microtype,subcaption}
\usepackage{scalerel}
\usepackage{graphicx}
\usepackage{epstopdf}
\usepackage{epsfig} 
\usepackage{pgfplots}
\usepackage{tikz,everypage}
\usetikzlibrary{arrows.meta}
\usetikzlibrary{arrows}
\usepackage[percent]{overpic}
\usepackage[]{units}
\usepackage{lipsum} 
\usepackage[dvipsnames]{xcolor}
\usepackage[onehalfspacing]{setspace}

\newcommand{\subscriptsize}{5pt}
\newcommand{\subscriptsizeAlt}{6pt}

\newcommand{\alphaSP}{\ensuremath{\alpha_{\scaleto{\text{SP}}{\subscriptsize}}}}
\newcommand{\betaSP}{\ensuremath{\beta_{\scaleto{\text{SP}}{\subscriptsize}}}}

\newcommand{\pBar}{\ensuremath{\bar{p}}}
\newcommand{\pA}{\ensuremath{p_{\scaleto{\text{A}}{\subscriptsize}}}}
\newcommand{\pB}{\ensuremath{p_{\scaleto{\text{B}}{\subscriptsize}}}}
\newcommand{\pC}{\ensuremath{p_{\scaleto{\text{C}}{\subscriptsize}}}}

\newcommand{\pASP}{\ensuremath{p_{\scaleto{\text{A,SP}}{\subscriptsizeAlt}}}}
\newcommand{\pBSP}{\ensuremath{p_{\scaleto{\text{B,SP}}{\subscriptsizeAlt}}}}
\newcommand{\pCSP}{\ensuremath{p_{\scaleto{\text{C,SP}}{\subscriptsizeAlt}}}}

\newcommand{\pABSP}{\ensuremath{p_{\scaleto{\text{AB,SP}}{\subscriptsizeAlt}}}}
\newcommand{\pBCSP}{\ensuremath{p_{\scaleto{\text{BC,SP}}{\subscriptsizeAlt}}}}

\newcommand{\uA}{\ensuremath{u_{\scaleto{\text{A}}{\subscriptsize}}}}
\newcommand{\uB}{\ensuremath{u_{\scaleto{\text{B}}{\subscriptsize}}}}
\newcommand{\uC}{\ensuremath{u_{\scaleto{\text{C}}{\subscriptsize}}}}

\newcommand\copyrighttext{%
	\footnotesize \textcopyright 2020 Frontiers.  Personal use of this material is permitted.  Permission from Frontiers must be obtained for all other uses, in any current or future media, including reprinting/republishing this material for advertising or promotional purposes, creating new collective works, for resale or redistribution to servers or lists, or reuse of any copyrighted component of this work in other works.}
\newcommand\copyrightnotice{%
	\begin{tikzpicture}[remember picture,overlay]
		\node[anchor=north,yshift=-10pt] at (current page.north) {\fbox{\parbox{\dimexpr\textwidth-\fboxsep-\fboxrule\relax}{\copyrighttext}}};
	\end{tikzpicture}%
}

\def\keyFont{\fontsize{8}{11}\helveticabold }
\def\firstAuthorLast{Hofer {et~al.}} 
\def\Authors{Matthias Hofer\,$^{1,*}$, Carmelo Sferrazza\,$^{1}$ and Raffaello D'Andrea$^{1}$\,}
\def\Address{$^{1}$Institute for Dynamic Systems and Control, ETH Zurich, Zurich, Switzerland}
\def\corrAuthor{Matthias Hofer, Institute for Dynamic Systems and Control, ETH Zurich, Sonneggstrasse 3, 8092 Zurich, Switzerland}

\def\corrEmail{hofermat@ethz.ch}

\begin{document}
\onecolumn
\firstpage{1}

\title[A Vision-based Sensing Approach for a Spherical Soft Robotic Arm]{A Vision-based Sensing Approach for a Spherical Soft Robotic Arm} 

\author[\firstAuthorLast ]{\Authors} 
\address{} 
\correspondance{} 

\extraAuth{}

\maketitle

\copyrightnotice



\begin{abstract}

Sensory feedback is essential for the control of soft robotic systems and to enable deployment in a variety of different tasks. Proprioception refers to sensing the robot's own state and is of crucial importance in order to deploy soft robotic systems outside of laboratory environments, i.e. where no external sensing, such as motion capture systems, is available.   

A vision-based sensing approach for a soft robotic arm made from fabric is presented, leveraging the high-resolution sensory feedback provided by cameras. No mechanical interaction between the sensor and the soft structure is required and consequently, the compliance of the soft system is preserved. The integration of a camera into an inflatable, fabric-based bellow actuator is discussed. Three actuators, each featuring an integrated camera, are used to control the spherical robotic arm and simultaneously provide sensory feedback of the two rotational degrees of freedom. A convolutional neural network architecture predicts the two angles describing the robot's orientation from the camera images. Ground truth data is provided by a motion capture system during the training phase of the supervised learning approach and its evaluation thereafter. 

The camera-based sensing approach is able to provide estimates of the orientation in real-time with an accuracy of about one degree. The reliability of the sensing approach is demonstrated by using the sensory feedback to control the orientation of the robotic arm in closed-loop.

\tiny
\keyFont{ \section{Keywords:} Soft robotics, proprioception, vision-based sensing, computer vision, supervised machine learning, fabric bellow, pneumatic actuation}

\end{abstract}

\section{Introduction}\label{sec:Introduction}

Soft robots show promise to overcome challenges encountered with rigid robots due to their versatility resulting from the soft materials employed \citep{PPolygerinos_SRR}. Their intrinsic mechanical properties are beneficial in terms of safety, allowing for close human-robot collaboration \citep{safety_soft_robots}. The academic relevance of the field is reflected by an increasing number of publications and growing attention within the field of robotics in general \citep{softRoboticsAcademic}. However, the potential of soft robots comes with several challenges, such as complex dynamics that are difficult to model and limit the application of open-loop control \citep{DRus_Review}. Therefore, sensory feedback is indispensable for accurate control and deployment in real-word applications \citep{HWang_TowardsPerceptiveSR}. 

A wide range of sensing principles are explored to provide proprioceptive feedback, i.e. feedback of the robot's own state. Vision-based approaches relying on internal cameras to observe the deformation of soft materials are promising because the sensor provides a high resolution and is not required to mechanically interact with the soft material that is observed. 

\begin{figure}[ht]
\center
\includegraphics[width=9.1cm]{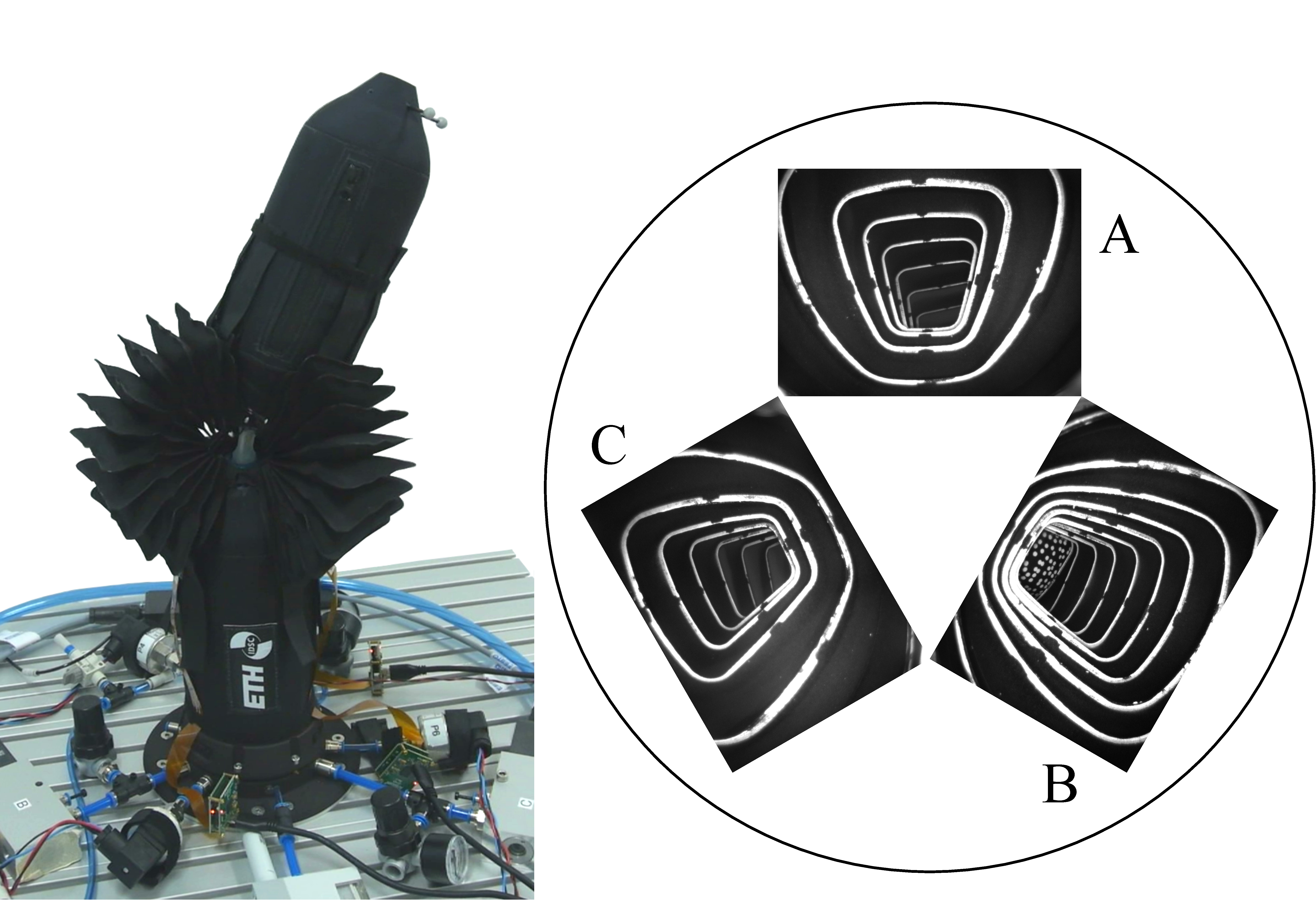}
\caption{The figure on the left shows the spherical robotic arm used for evaluation of the vision-based sensing approach proposed in this work. The figure on the right shows images from the cameras placed inside three inflatable bellow actuators. The arrangement of the camera images matches a view from the bottom looking upwards. The orientation of the movable link can be observed in certain actuator elongations and deformations that are observed by the internal cameras.}\label{fig:titleImage}
\end{figure}

Our method originates from the work documented in \cite{werner2019visionbased}. The principle idea is to integrate a camera into a fabric-based bellow actuator and use three of these actuators to control a spherical robotic arm (see Figure \ref{fig:titleImage}). The two rotational degrees of freedom of the robotic arm are estimated from the actuator expansion and deformation observed by the three internal cameras. 
The mapping from camera images to the orientation of the movable link is identified by relying on a supervised learning approach with ground truth information provided by a motion capture system. A convolutional neural network architecture maps the camera images to the orientation of the robot arm. The sensing pipeline developed is deployed in real-time and used for closed-loop control of the robotic arm.

\vspace{6pt}
\subsection{Related Work}
A number of different approaches are investigated for retrieving the shape of a soft robot based on internal sensors only \citep{HWang_TowardsPerceptiveSR}. A common approach is to combine sensing algorithms with machine learning techniques to retrieve the quantity of interest from the raw sensor output. An overview of such applications for sensing and control is provided in \cite{MLsoftRobotics}. Any sensor relies on the change of a physical property induced by the soft structure undergoing a deformation or expansion. Different sensing principles and the applications thereof are summarized below.

Resistive and piezoresistive strain sensors detect a change in resistance caused by material deformation \citep{resistiveSensors}. The advantage of resistive and piezoresistive sensors is their relatively easy fabrication and integration \citep{HWang_TowardsPerceptiveSR}. An application of a resistive sensor is presented in \cite{Thuruthel_softSensor}, in combination with a recurrent neural network that maps the raw sensor output to the bending state of a soft finger. Piezoresistive sensors are employed in \cite{ryanProprioception} to form a soft, proprioceptive sensor skin, which can be attached to a soft robotic system. A recurrent neural network predicts the robot configuration based on the sensor measurements.

A capacitive strain sensor is presented in \cite{capacitive} and deployed for an intelligent glove application. An approach based on the change of inductance is presented in \cite{inductanceSensing} for a bellows-driven continuum robot and used in closed-loop for feedback control. A sensing approach relying on a magnet and a Hall sensor integrated into a soft bending actuator is documented in \cite{magnetic}. The relative orientation between the magnet and sensor varies as the soft structure deforms, causing the observed magnetic field to change. The sensing type is simple to integrate and can be used to control the bending angle of the actuator. The method presented in \cite{accousticSensing} leverages an acoustic sensing principle. A speaker and microphone are integrated into a soft extensible pneumatic actuator and used to detect the changing resonance characteristics as the elongation of the actuator varies. The sensing approach is used in closed-loop to track the desired length of the actuator. The work presented in \cite{BionicHandAss2011} relies on commercially readily available Bowden cable potentiometers to retrieve and control the shape of a continuum robot arm.

Optical sensors detect changes in the light transmission of a soft medium when deformed. A common approach is to measure the varying light intensity. The integration of macrobend stretch sensors into a soft arm is documented in \cite{opticalFibreInte}. The bending of the light transmitting fiber causes the intensity of the light transmission to change. The use of stretchable optical waveguides is reported in \cite{Zhaoeaai7529} to provide sensing capabilities for shape and force for a prosthetic hand. Closed-loop control is also demonstrated in \cite{ofs_control}. A fabric-based bellow actuator, similar to ours, is used in \cite{HYang_ModelAndAnalysis} as the light reflecting surface. A photo transistor is attached to one end of the linear bellow actuator and measures the light intensity from a light emitting diode (LED) attached to the opposing end of the actuator. The intensity decreases as a function of the actuator elongation. The advantage of optical sensors is the high sensitivity and repeatability \citep{ZKappassov_TactileSensing} and the fact that the electronics can be placed outside of the sensing area \citep{HWang_TowardsPerceptiveSR}. 

Another optical sensing principle relies on fiber Bragg grating. The use of a distributed fiber Bragg sensor network is demonstrated in \cite{FBG_Pfeifer} for a cone-shaped soft manipulator made from silicone.

The discussion of camera-based sensing approaches is limited here to examples relying on internal cameras. Methods purely based on external cameras, including motion capture systems, are not discussed. Cameras visually observe material deformation through the movement of visual features located in or attached to the soft material. Camera-based sensing is actively explored in the field of tactile sensing with an overview provided in \cite{tactileSensing}. 

Compared to placing the cameras externally, an advantage of integrating cameras into the soft system and pointing them to the interior of the structure, is the possibility to design the area observed by the camera for best performance without external influence. The application of a pattern to the interior surface of the structure allows for the provision of rich information about the deformation state and control of the lighting conditions. Consequently, the sensing approach does not depend on the visual features present in the environment or the existing external lighting conditions.

A vision-based tactile sensor including pneumatic actuation is presented in \cite{BMcInroe_TowardsAS}. A combination of blob detection and optical flow is used to track a number of markers and infer contact conditions and membrane shear. Increasing the internal pressure allows for inflation of the membrane and thereby control of the interaction force. A tactile sensor named TacEA combines vision-based tactile sensing, pneumatic actuation and electroadhesive grasping capabilities and is presented in \cite{Xiang_2019}. The sensing principle relies on the TacTip family as presented in \cite{TacTip}. After an object is gripped using the electroadhesion, releasing the object can take a considerable amount of time. Pneumatic actuation, i.e. inflation of the soft membrane, allows the object to be released quickly. Other camera-based tactile sensors are presented in \cite{WYuan_GelSight} and \cite{CSferrazza_DesignMot}. 

A method to sense the three-dimensional shape of a soft robot relying on a self-observing camera is documented in \cite{DeepVision}. External depth cameras provide ground truth to train a neural network, which predicts the shape of the object only from images of the self-observing cameras. The approach is executed on a graphics processing unit (GPU) and provides the three-dimensional deformation of soft objects in real-time. A vision-based sensing approach providing both proprioceptive and exteroceptive sensing is demonstrated in \cite{exoskeleton} for an exoskeleton-covered soft finger. The sensing method relies on a convolutional neural network architecture being executed on a GPU that is able to predict the shape of a single finger in real-time and to classify objects which are grasped with a gripper made from two fingers. 

In \cite{OLIVEIRA2020}, a sensing method is presented to measure the bending deformation of a soft link and detect interactions with the environment. A camera is mounted inside an inflatable and compliant link. A blob detection algorithm relates the two-dimensional tip position displacement to the location of a center blob and changes in the relative positions of lateral blobs are interpreted as a contact with the environment. The compliant link is actuated by two electric motors and a filtering approach is employed to the input signals to reduce an excitation of the lowest natural frequency of the link. The internal pressure is increased and thereby demonstrated to compensate for a shift in the lowest natural frequency when a payload is attached to the link.

Sensing approaches relying on a camera are promising because the sensor (i.e. the camera) is not required to mechanically interact with the soft material being observed. Therefore, the compliance of the sensor and the soft material are not required to match, simplifying material selection and avoiding stress concentrations at the interface between the sensor and the soft material, which otherwise can limit the maximum number of load cycles of the soft robotic system. Furthermore, cameras provide a high resolution, they are not affected by environmental influences such as temperature or electromagnetic noise and their low cost enables the deployment of multiple sensors in soft robotic systems. The challenge with cameras is to integrate the rigid sensor into the soft structure. The size of the camera itself can impose design constraints and the material deformation of interest is required to lie in the visible area of the camera, which can further complicate integration. Additionally, the high-dimensional sensor output needs to be processed in real-time, which requires computational capacity \citep{ZKappassov_TactileSensing}.

\vspace{6pt}
\subsection{Contribution}
While camera-based sensing approaches have been demonstrated for a soft plush robot, soft fingers and a compliant link, we demonstrate a vision-based sensing approach for a fabric-based bellow actuator used in a soft robotic arm. Our approach relies on the integration of a small camera with a footprint of $\unit[7]{}$x$\unit[7]{mm}$ and a distinctive white pattern which is applied to the interior surface of the actuator. Multiple LEDs are integrated to control the illumination. 

A convolutional neural network architecture is trained and used to map the raw camera images to the rotational degrees of freedom of the robotic arm. We show that a lightweight network architecture, which can be deployed on a regular laptop computer without GPU support, can predict the orientation of the robot arm at $\unit[30]{Hz}$ and achieves a root-mean-square accuracy of $\unit[1.25]{^\circ}$.

While camera-based interaction force control is demonstrated in \cite{BMcInroe_TowardsAS} and feed-forward vibration control in \cite{OLIVEIRA2020}, no closed-loop position control relying on feedback from cameras has been demonstrated for a soft robotic system. We extend the results presented in \cite{werner2019visionbased} for a single, linear actuator, to the control of a spherical robotic arm using three actuators each including an internal camera.

\vspace{6pt}
\subsection{Outline}
The remainder of this paper is organized as follows: Section \ref{sec:Hardware} presents the design of the soft bellow actuator and the integration of the camera and the peripherals required. The machine learning pipeline to retrieve the orientation from the camera images is discussed in Section \ref{sec:CameraBasedSensing} and the control approach employed in Section \ref{sec:Control}. Results showing the real-time prediction capability of the sensing approach are presented in Section \ref{sec:Results}, along with a validation of the sensing approach to provide feedback for closed-loop control experiments. Finally, a conclusion is drawn in Section \ref{sec:Discussion}.
\vspace{14pt}

\section{Materials and Methods}\label{sec:MaterialsAndMethods}
The hardware used for realizing the camera-based sensing approach is discussed in the first part of this section. In a second part, a supervised machine learning approach is presented that maps the camera images to the angles describing the orientation of the robotic arm. The section is concluded with a brief description of the controller employed on the robotic arm.
\vspace{6pt}
\subsection{Hardware}\label{sec:Hardware}
We start with an overview of the spherical robotic arm used for evaluation of the camera-based sensing method. The following sections outline principle design considerations regarding the vision-based actuator, the manufacturing of the soft actuators and the required camera peripherals employed. The section is concluded by a discussion of the integration of the camera into the actuator. 

\vspace{8pt}
\subsubsection{Spherical Robotic Arm}
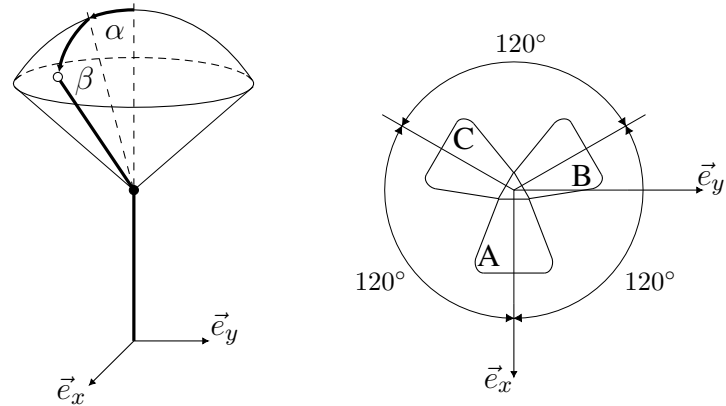
\begin{figure}[h]
	\centering


\usetikzlibrary{arrows.meta}
\newcommand{\arr}{-{Latex[length=1mm, width=1mm]}}
\newcommand{\arrd}{{Latex[length=1mm, width=1mm]}-{Latex[length=1mm, width=1mm]}}

\begin{tikzpicture}[scale = 1]
\draw[very thick] (0,0) -- (0,2);
\draw[very thick] (0,2) -- (-0.97,3.45);
\draw[dashed] (0,2.05) -- (0,4.52);
\draw[dashed] (-0.005,2.01) -- (-0.62,4.38);
\draw [\arr, very thick] (0,4.39) arc [radius=1.6, start angle=90, end angle= 112];
\draw [\arr, very thick] (-0.591,4.272) arc [radius=1.25, start angle=132, end angle= 170];
\draw[\arr] (0,0) -- (1,0);
\draw[\arr] (0,0) -- (-0.6,-0.6);
\node at (-0.82,-0.65) {$\vec{e}_x$};
\node at (1.2,0.17) {$\vec{e}_y$};
\node at (-0.25,4.1) {$\alpha$};
\node at (-0.65,3.45) {$\beta$};
\draw [fill] (0,2) circle [radius=0.06];
\draw [] (-1.0,3.5) circle [radius=0.06];

\coordinate (O) at (0,2);

\draw[ultra thin] (0,2) to [edge label = $$] (-1.57,3.37);
\draw[ultra thin] (0,2) -- (1.57,3.37);

\draw[] (-1.57,3.46) arc [start angle = 140, end angle = 40,
  x radius = 20.4mm, y radius = 26mm];
\draw[densely dashed] (-1.559,3.46) arc [start angle = 170, end angle = 10,
  x radius = 15.8mm, y radius = 3.6mm];
\draw[] (-1.489,3.52) arc [start angle=-200, end angle = 20,
  x radius = 15.8mm, y radius = 3.15mm];

\begin{scope}[shift={(1,0)}]
\draw[\arr] (5.5,2) -- (6.5,2);
\draw[\arr] (4,0.5) -- (4,-0.5);
\draw[ultra thin] (5.5,2) -- (4,2) -- (4,0.5);
\node at (3.8,-0.5) {$\vec{e}_x$};
\node at (6.6,2.15) {$\vec{e}_y$};

\begin{scope}[rotate around = {30:(4,2)}]
\draw[ultra thin] (4,2) -- (6,2);
\end{scope}
\begin{scope}[rotate around = {150:(4,2)}]
\draw[ultra thin] (4,2) -- (6,2);
\end{scope}
\draw [\arrd,ultra thin,domain=30:150] plot ({4+1.7*cos(\x)}, {2+1.7*sin(\x)});
\node at (4.1,3.95) {\tiny $120^{\circ}$};
\draw [\arrd,ultra thin,domain=150:270] plot ({4+1.7*cos(\x)}, {2+1.7*sin(\x)});
\node at (2.25,0.8) {\tiny $120^{\circ}$};
\draw [\arrd,ultra thin,domain=270:390] plot ({4+1.7*cos(\x)}, {2+1.7*sin(\x)});
\node at (5.8,0.8) {\tiny $120^{\circ}$};

\draw (3.5,1.1) -- (3.8,1.88) -- (4.2,1.88) -- (4.5,1.1);
\draw (3.6,0.9) -- (4.4,0.9);
\draw[] (4.5,1.1) to [out=-70,in=0] (4.4,0.9);
\draw[] (3.5,1.1) to [out=-110,in=180] (3.6,0.9);
\node at (3.67,1.13) {A};

\begin{scope}[rotate around = {120:(4,2)}]
\draw (3.5,1.1) -- (3.8,1.88) -- (4.2,1.88) -- (4.5,1.1);
\draw (3.6,0.9) -- (4.4,0.9);
\draw[] (4.5,1.1) to [out=-70,in=0] (4.4,0.9);
\draw[] (3.5,1.1) to [out=-110,in=180] (3.6,0.9);
\end{scope}
\node at (4.9,2.2) {B};

\begin{scope}[rotate around = {240:(4,2)}]
\draw (3.5,1.1) -- (3.8,1.88) -- (4.2,1.88) -- (4.5,1.1);
\draw (3.6,0.9) -- (4.4,0.9);
\draw[] (4.5,1.1) to [out=-70,in=0] (4.4,0.9);
\draw[] (3.5,1.1) to [out=-110,in=180] (3.6,0.9);
\end{scope}
\node at (3.34,2.69) {C};
\end{scope}
\end{tikzpicture}
	\caption{The orientation parametrization of the spherical robotic arm is shown in the left hand plot. The static link is aligned with the inertial $z$-axis. A positive rotation of the movable link around the inertial $x$-axis is denoted by $\alpha$ and a positive rotation around the inertial $y$-axis is denoted by $\beta$. The top view of the actuator configuration in the inertial coordinate frame is shown in the right hand plot. The three actuators are arranged symmetrically around the inertial $z$-axis, where the actuator A is aligned with the inertial $x$-axis.}
	\label{fig:angleConvention}
\end{figure}

The spherical robotic arm is closely related to the system presented in \cite{RoboSoft2021} and consists of two inflatable links and three fabric-based bellow actuators that are arranged symmetrically around a soft silicone joint connecting the two links. The robot arm has two rotational degrees of freedom, which are described by the extrinsic Euler angles $\alpha$ and $\beta$ (see Figure \ref{fig:angleConvention}). The orientation of the movable link can be adjusted by inflating the actuators A, B and C to different pressures $p_{\text{A}}, p_{\text{B}}$ and $p_{\text{C}}$ to control the elongation of each actuator. Therefore, the three actuator pressures form the control inputs to the system. Note that each bellow actuator can not only expand longitudinally (when pressurized), but allows also for lateral deformation when the other actuators expand.
Since we have three control inputs for only two rotational degrees of freedom, it is also possible to control the stiffness of the joint. An intuitive way to understand this property is the fact that a certain orientation of the movable link can be attained by multiple pressure combinations, where the sole difference lies in the resulting joint stiffness. In this work, the capability of adjusting the joint stiffness is not explored and the reader is referred to \cite{HOFER2020102369} and \cite{RoboSoft2021} for more details.

\vspace{8pt}
\subsubsection{Vision-based Actuator}\label{sec:visionBasedActuator}
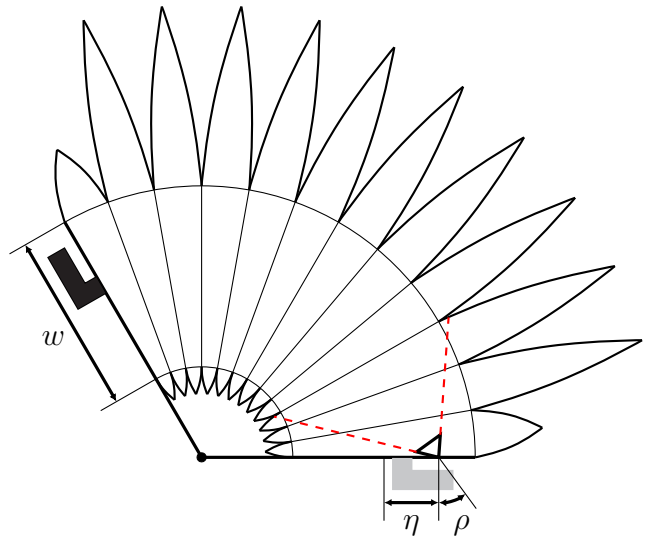
\begin{figure}
	\center
\begin{tikzpicture}[scale = 0.6]

\def\shift{2}
\def\openingAngleMin{-55}
\def\openingAngleMax{65}
\def\innerRadius{2}
\def\outerRadius{6}

\usetikzlibrary{arrows.meta}
\newcommand{\arr}{-{Latex[length=1mm, width=1mm]}}
\newcommand{\arrd}{{Latex[length=1mm, width=1mm]}-{Latex[length=1mm, width=1mm]}}

\begin{scope}[rotate around = {55:(0,0)}]

\begin{scope}[rotate around = {\openingAngleMax:(0,0)}]
\draw[very thick] (0,0) -- (\outerRadius,0);
\draw[thick] (4.2,0) -- (4.2,0.7) -- (5.5,0.7) -- (5.5,0.3) -- (4.6,0.3) -- (4.6,0);
\fill[black] (4.2,0) -- (4.2,0.7) -- (5.5,0.7) -- (5.5,0.3) -- (4.6,0.3) -- (4.6,0) -- cycle;
\end{scope}

\begin{scope}[rotate around = {\openingAngleMin:(0,0)}]
\draw[very thick] (0,0) -- (\outerRadius,0);
\draw[thick, gray] (4.2,0) -- (4.2,-0.7) -- (5.5,-0.7) -- (5.5,-0.3) -- (4.6,-0.3) -- (4.6,0);
\fill[gray] (4.2,0) -- (4.2,-0.7) -- (5.5,-0.7) -- (5.5,-0.3) -- (4.6,-0.3) -- (4.6,0) -- cycle;
\end{scope}

\draw [fill] (0,0) circle [radius=0.1];

\draw[] (\openingAngleMin:\innerRadius) arc(\openingAngleMin:\openingAngleMax:\innerRadius);
\draw[] (\openingAngleMin:\outerRadius) arc(\openingAngleMin:\openingAngleMax:\outerRadius);

\foreach \angle in {-55,-45,-35,-25,-15,-5,5,15,25,35,45,55,65} 
{
	\begin{scope}[rotate around = {\angle:(0,0)}]
	\draw (\innerRadius,0) -- (\outerRadius,0);
	\end{scope}
}

\foreach \angle in {-40,-30,-20,-10,0,10,20,30,40,50} 
{
	\begin{scope}[rotate around = {\angle:(0,0)}]
	\draw[thick] plot [smooth, tension=0.9 ] coordinates {(5:\outerRadius) (8,0.4) (10,0)};
	\draw[thick] plot [smooth, tension=0.9 ] coordinates {(-5:\outerRadius) (8,-0.4) (10,0)};
	
	\draw[thick] plot [smooth, tension=0.9 ] coordinates {(5:\innerRadius) (1.7,0.12) (1.4,0)};
	\draw[thick] plot [smooth, tension=0.9 ] coordinates {(-5:\innerRadius) (1.7,-0.12) (1.4,0)};
	\end{scope}
}

\foreach \angle in {-50,60} 
{
	\begin{scope}[rotate around = {\angle:(0,0)}]
	\draw[thick] plot [smooth, tension=0.9 ] coordinates {(5:\outerRadius) (6.8,0.35) (7.5,0)};
	\draw[thick] plot [smooth, tension=0.9 ] coordinates {(-5:\outerRadius) (6.8,-0.35) (7.5,0)};
	
	\draw[thick] plot [smooth, tension=0.9 ] coordinates {(5:\innerRadius) (1.7,0.12) (1.4,0)};
	\draw[thick] plot [smooth, tension=0.9 ] coordinates {(-5:\innerRadius) (1.7,-0.12) (1.4,0)};
	\end{scope}
}

\begin{scope}[rotate around = {\openingAngleMin:(0,0)}, shift={(5.2,0)}, rotate around = {36:(0,0)}]
\draw[very thick] (0,0) -- (130:0.5) -- (50:0.5) --(0,0);
\draw[] (0,0) -- (0, -1.4);
\draw[thick, red, dashed] (130:0.5) -- (130:3.72);
\draw[thick, red, dashed] (50:0.5) -- (50:3.1);
\end{scope}

\begin{scope}[rotate around = {\openingAngleMin:(0,0)}, shift={(5.2,0)}]
\draw[] (0,0) -- (0, -1.4);
\end{scope}

\begin{scope}[rotate around = {\openingAngleMin:(0,0)}, shift={(4,0)}]
\draw[] (0,0) -- (0, -1.4);
\end{scope}

\begin{scope}[rotate around = {\openingAngleMin:(0,0)}]
\draw[very thick, \arrd] (4,-1) -- (5.2, -1);
\draw[very thick,\arrd] (5.2,-1) arc [radius=1, start angle=-90, end angle= -54];
\node at (4.6,-1.5) {$\eta$};
\node at (5.7,-1.5) {$\rho$};
\end{scope}

\begin{scope}[rotate around = {\openingAngleMax:(0,0)}]
\draw[very thick, \arrd] (\innerRadius,1) -- (\outerRadius, 1);
\draw[] (\innerRadius,0) -- (\innerRadius, 1.4);
\draw[] (\outerRadius,0) -- (\outerRadius, 1.4);
\node at (3.9,1.5) {$w$};
\end{scope}


\end{scope}

\end{tikzpicture}
	\caption{The figure shows a simplified sketch of the cross section of an actuator with the visible area of the camera indicated in red (dashed). Angle connectors are attached to both the top side of the actuator (shown in black in the left of the figure) and bottom (shown in gray in the bottom of the figure). The bottom connector is used to pressurize the actuator and the top connector to align it with the movable link. The inner opening which connects neighboring cushions has a width denoted by $w$ and plays a crucial role in the resulting visible area of the camera. If the opening is sufficiently wide, the majority of the cushions are within the visible area of the camera. The area of the actuator deformation covered by the camera is increased, if the camera is placed with an offset $\eta$ with respect to the center of the inner opening and tilted by an angle $\rho$ with respect to the normal direction.}
	\label{fig:designConsiderations}
\end{figure}

Principle design considerations for the vision-based actuator are discussed in this section. The fabric-based actuator consists of individual cushions, which are combined at seams around an inner opening, and forming a bellow-type actuator. A simplified sketch of the actuator is shown in Figure \ref{fig:designConsiderations}. The actuator combines soft actuation when pressurized and sensing of the actuator's angular elongation and lateral deformation through an integrated camera. The combined actuation and sensing system needs to address several requirements for a successful deployment. These requirements are discussed below.

The camera field of view should cover a large range of the actuator expansion to provide sensory feedback over a large range of the movable link. The sensitivity of the sensing approach is maximized if the actuator expansion and deformation cause large variations in the camera images observed. Increasing the width of the inner opening clearly improves the visible area of the camera. Placing the camera with an offset, $\eta$, towards the outer edge of the actuator and tilting the camera by an angle, $\rho$, towards the center of the actuator increases the visible area of the actuator deformation covered by the camera (compare Figure \ref{fig:designConsiderations}).

The angular expansion of the entire actuator should be maximized such that the angular range of the movable link is maximized. Therefore, the angular expansion of a single cushion should be maximized by either increasing the radial width of the actuator, which is done for all cushions except the top and bottom cushions, or reducing the width of the inner opening which violates the previously discussed design requirement. The angular expansion of the actuator can further be improved by increasing the number of cushions employed, which however also leads to a higher production time.  

Finally, the actuator needs to be compatible with the links of the robotic arm. Therefore, the ratio of linear and angular expansion of the actuator need to approximately fit the robotic arm. The ratio of angular and linear expansion mainly depend on the ratio between inner opening width to the radial width of the actuators. Additionally, the location of the inner opening also affects the ratio between angular and linear expansion, where a central positioning yields a linear actuator and a placement of the opening off-center primarily leads to an angular expansion of the actuator. 

The final actuator geometry addressing all design requirements mentioned is detailed in the supplementary files provided.

\vspace{6pt}
\subsubsection{Manufacturing of the Soft Actuator}\label{subsec:ManufacturingOfTheSoftActuator}
After defining the requirements of the vision-based actuator in the previous section, the manufacture of the inflatable bellow actuators is discussed here. The manufacture of the rotary actuator is similar to the design presented in \cite{werner2019visionbased} for a linear actuator. The fabrication method as presented in \cite{yang2018new} is applied. The actuators are made from fabric sheets consisting of a sandwich structure. A layer of thermoplastic polyurethane (TPU) film (HM65-PA, 0.1 mm by perfectex) is used inbetween two layers of poplin fabric (polyester cotton blend 65/35 by extremtextil) fused in a heat press (TS7 swingaway heat press by Secabo). The resulting fabric material is inextensible, airtight and sturdier than the single layers of poplin fabric.

The actuator is composed of twelve single cushions, with each cushion being constructed from two pieces of fabric. The pieces have a cutout in the middle (except for the top and bottom part) where the individual cushions are connected to form a bellow actuator. As mentioned in the previous section, placing the cutout off-center results in a rotary expansion type. Additional TPU ring-shaped seam pieces are prepared to combine the fabric pieces. The actuator is built by stacking the fabric and TPU pieces and fusing them sequentially in a bottom-up process. All fabric and TPU pieces are prepared with a laser cutter and a detailed description of the fabrication procedure can be found in \cite{yang2018new} (Layered Manufacturing-Type I).

Before the fabric layers are fused together, a white pattern is applied to the layers facing the camera to provide visual features with a high contrast to the black fabric. The pattern is cut from adhesive stencil film (S380 by ASLAN) and applied with textile spray paint (319921 textile spray paint by DupliColor) in consecutively applied thin layers. The pattern includes dots with a diameter of $\unit[2]{mm}$ on the top layer and rings around the opening of the middle cushions.

\vspace{8pt}
Since the fabric material is opaque, a light source is required to illuminate the interior of the actuator and make the white pattern visible to the camera. The camera electronics, including its peripherals, are discussed the next section. The files of all fabric parts are provided in the supplementary files of this publication.

\vspace{10pt}
\subsubsection{Camera Electronics Setup}

The camera electronics setup used in each actuator is depicted in Figure \ref{fig:design_combo} (A). The camera (OV7251 by OmniVision) houses a CMOS VGA sensor with a maximum resolution of 640x480 pixels and a corresponding frame rate of 100 frames per second. The camera has a footprint of $\unit[7]{}$\!x$\unit[7]{mm}$ and is significantly smaller than the camera employed in \cite{werner2019visionbased}, therefore simplifying integration. The camera is connected to an adapter board which reroutes the pins to an Arducam USB Camera Shield (UC-425 Rev. C). A custom LED board is powered and controlled via the adapter board. The light intensity is adjusted by setting the duty cycle of a pulse width-modulated signal. A constant duty cycle of $\unit[0.22]{}$ is used throughout this work. The camera employed allows synchronization of the cameras over pins routed to the adapter board and connected between the three cameras. The camera of actuator A takes the role of the leading camera that triggers the picture and the other two cameras take the role of followers. The Arducam USB Camera Shield SDK library is used for software integration. The schematics of the LED board and the adapter board are provided in the supplementary files of this publication.
\vspace{10pt}
\subsubsection{Camera Integration}
\begin{figure}
	\center
	\includegraphics[width=13cm]{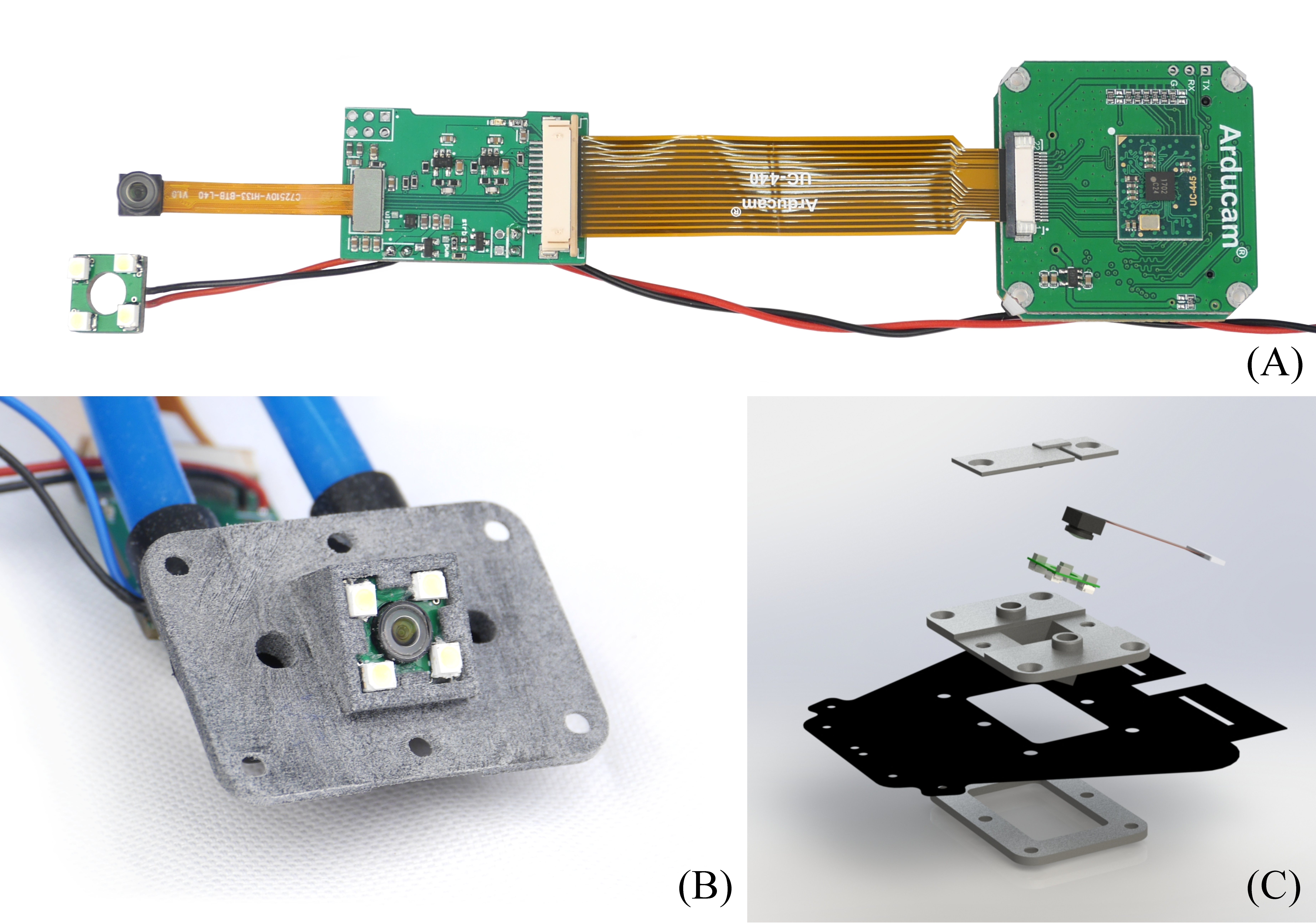}
	\caption{(A) The figure shows the camera electronics used in each actuator. The camera and LED board are connected to a to a custom-made adapter board which reroutes the camera pins and powers the LED board. The adapter board includes pins which are used for synchronizing multiple cameras and is powered over the black/red cables. The adapter board is connected to an Arducam USB Camera Shield (UC-425 Rev. C) with USB interface (USB cable not shown). (B) The picture shows the front view of the camera adapter housing the camera and the enclosing LED board. The camera is tilted by an angle of $\unit[25]{^\circ}$ with respect to the normal direction of the adapter plane. The pressure is measured and controlled over the blue tubing connected to the adapter over black angle connectors and routed to the two openings next to the camera. (C) The figure shows a rendering of the camera interface. The bottom fabric layer is sandwiched between the camera adapter and a 3D printed flange ring, which are fastened by six screws. The LED board and camera are inserted into the adapter from the top and the camera cable is routed through a slit in the top piece. Silicone glue is used to seal all interfaces.}
	\label{fig:design_combo}
\end{figure}

\begin{figure}
	\center
	\includegraphics[width=15cm]{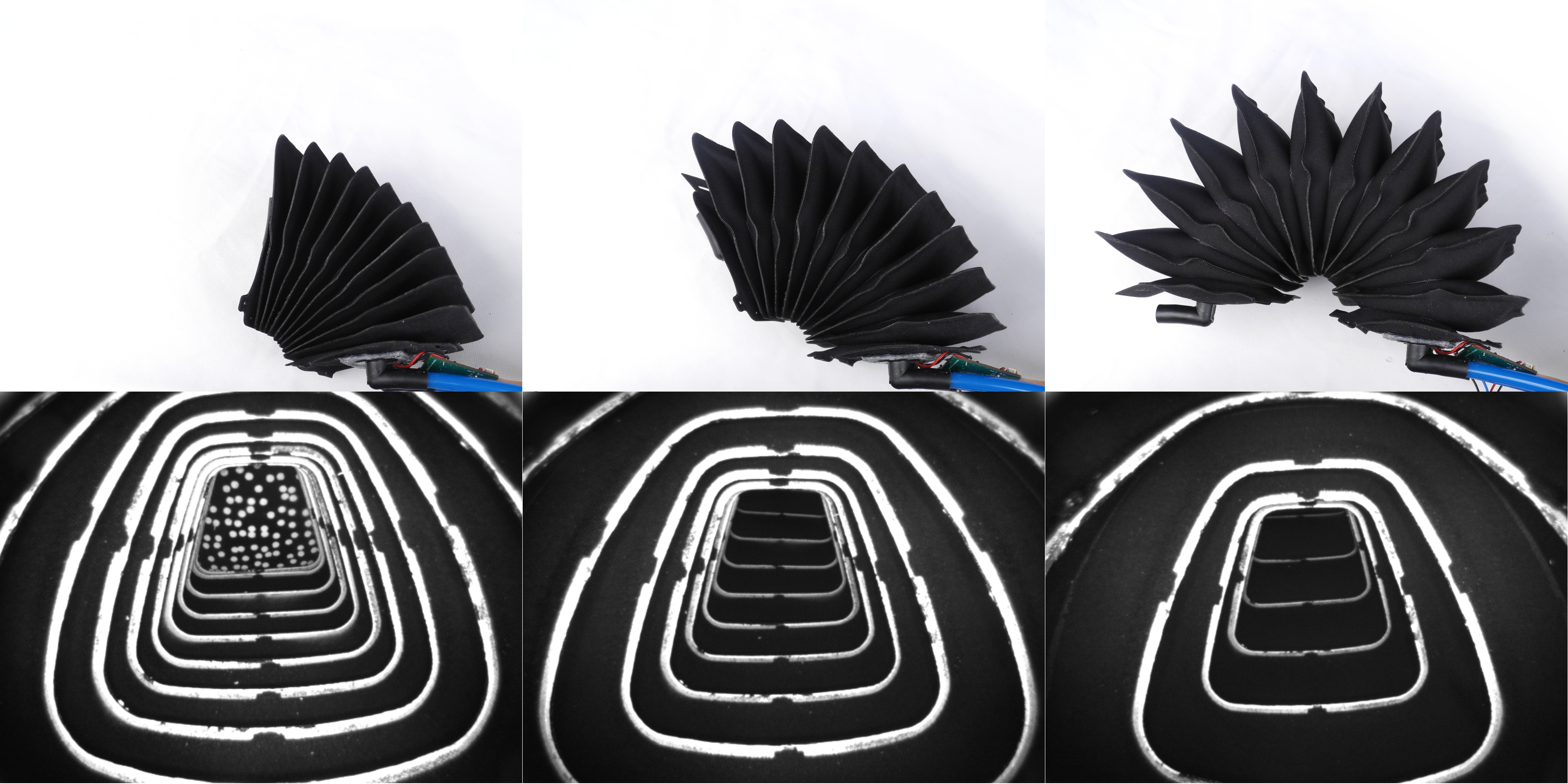}
	\caption{The picture shows a single actuator inflated to different expansions and the corresponding image from the internal camera. The number of cushion rings visible to the internal camera decreases as the actuator expands. The light intensity is set such that the white pattern is visible over the full range of the actuator expansion.}
	\label{fig:cameraActuator}
\end{figure}

In this section we discuss the integration of the camera and its peripherals into the soft actuator. Only the camera and the LED board are mounted inside the actuator. The other peripherals shown in Figure \ref{fig:design_combo} (A) are placed outside the actuator. The camera and the LED board should point in the same direction irrespective of the actuator expansion. Therefore, both the camera and the LED board are glued (Silicone multi-purpose sealant 732 by Dow Corning) to the 3D printed adapter piece (made from PA12). The camera offset $\eta$ and angle $\rho$, as discussed in Section \ref{sec:CameraBasedSensing}, can be addressed in the design of the 3D printed adapter. The adapter is fixed to an opening in the bottom layer of the actuator over a flange-like structure (see Figure \ref{fig:design_combo} (C)). The CAD files of the camera adapter are provided in the supplementary files of this publication.
The resulting actuator is shown in Figure \ref{fig:cameraActuator}, when inflated to different elongations with the image from the internal camera alongside. Although the camera adapter is made from rigid material, the adapter is enclosed by the actuator and the static link parts shielding the rigid part towards the surroundings. The camera electronics could be routed internally at the cost of a higher design complexity.
\vspace{6pt}
\subsection{Camera-based Sensing}\label{sec:CameraBasedSensing}
The method to predict the orientation of the robotic arm from the internal cameras is discussed in this section. The identification of the mapping from camera images to the angles describing the orientation of the robotic arm is posed as a supervised learning problem with ground truth data available by means of a motion capture system. The approach presented in \cite{werner2019visionbased} relies on a feature engineering step, followed by a support vector regression to predict the actuator deformation. The key advantage is the limited training complexity of the support vector regression, which came at the cost of the required feature engineering step. In this work, a lightweight convolutional network is used to predict the orientation of the robot arm from the camera images. The end-to-end learning approach bypasses any feature engineering step, but requires more training data resulting in longer training times. The network architecture is outlined in the first part of this section. The data collection and model learning are discussed thereafter.

\vspace{6pt}
\subsubsection{Data collection}
Ground truth data is provided by an infrared motion capture system running at $\unit[200]{Hz}$ and providing sub-millimeter accuracy of the $x$-$y$-$z$-position of the tip of the movable link. The extrinsic Euler angles $\alpha$ and $\beta$ are calculated by applying the following formulas,
\begin{linenomath}
\begin{align}
	\alpha& = \text{arcsin} (-(y-y_0)/R)\\
	\beta & = \text{arcsin} ((x-x_0)/(R\cdot\text{cos}(\alpha))).
\end{align}
\end{linenomath}
Thereby, $(x_0, y_0)$ denotes the coordinates of the pivot point and $R$ the radius of the movable link, which are both determined in a calibration process. 

The data collection includes storing the images from the three internal cameras of the current link orientation and the corresponding ground truth labels $\alpha$ and $\beta$. In order to cover the $\alpha$-$\beta$ plane uniformly, a position controller as discussed in Section \ref{sec:Control} is used to track a regular grid of $\alpha$ and $\beta$ set points in the range of $[-\unit[30]{},\unit[30]{^\circ}]$. The camera images and the corresponding ground truth labels are recorded at a rate of $\unit[10]{Hz}$.  

The data is preprocessed by first sub-sampling each image using linear interpolation to a resolution of 120x160 pixels. All pixel values are converted to floating point format and normalized to the interval $[-1,1]$. A training data set of approximately $\unit[54000]{}$ images and labels is collected (corresponding to $\unit[90]{min}$ of data) and a validation data set of approximately $\unit[15000]{}$ images and labels is recorded (corresponding to $\unit[25]{min}$ of data). 

\vspace{20pt}
\subsubsection{Network architecture}

The network architecture used in this work is related to LeNet as documented in \cite{LeNet}. The main building block is a convolutional layer followed by a nonlinear activation (i.e. ReLU) and a max pooling step. This building block is repeated three times, before the output of the last pooling step is fed into two fully connected layers predicting the two dimensional output. The resulting network is required to provide inference in real-time on a standard laptop computer without GPU support. Therefore, the maximum size of the network is limited. The following network exhibits a good trade-off between prediction accuracy and computational complexity.

\begin{figure}
	\centering
	\includegraphics[width=15cm]{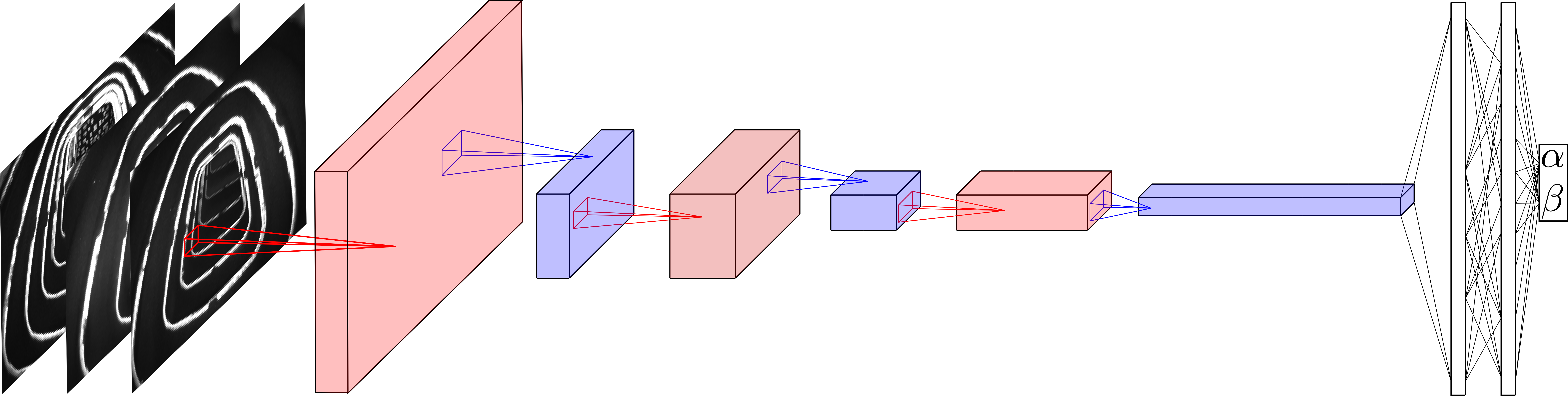}
	\caption{The figure depicts the architecture of the neural network employed. The input consists of the three camera images sub-sampled to a resolution of 120x160 pixels. First a convolutional layer (in red) is applied, with four output channels followed by a nonlinear activation (not shown) and a max pooling step (in blue) reducing the size of the image to 24x32. The procedure is repeated twice more, while the number of channels is increased to eight and 16, respectively. The pooling steps reduce the size to 6x8 after the second and to 3x4 after the third max pooling step. Finally, two fully connected layers are applied which output the angles $\alpha$ and $\beta$. The pooling layers reduce the number of parameters and consequently the computational complexity significantly, while retaining the most important features.}
	\label{fig:neuralNet}
\end{figure}

All convolutional layers have a kernel of size three, a stride of one and a padding of one. The max pooling kernel sizes (and the corresponding strides) are chosen as $(5, 4, 2)$ in the first, second and third layers, respectively. No padding is used for the pooling step. The output of the last pooling layer is fed into two fully connected layers with 40 neurons each and ReLU activation functions. The network architecture is depicted in Figure \ref{fig:neuralNet}. The network has a total number of 9378 parameters, which is about half of LeNet-4 with roughly 17000 parameters. 
\graphicspath{{figs/}}

\vspace{20pt}
\subsubsection{Model Learning}\label{sec:modelLearning}
\begin{figure}
	\center
	\includegraphics[width=10cm]{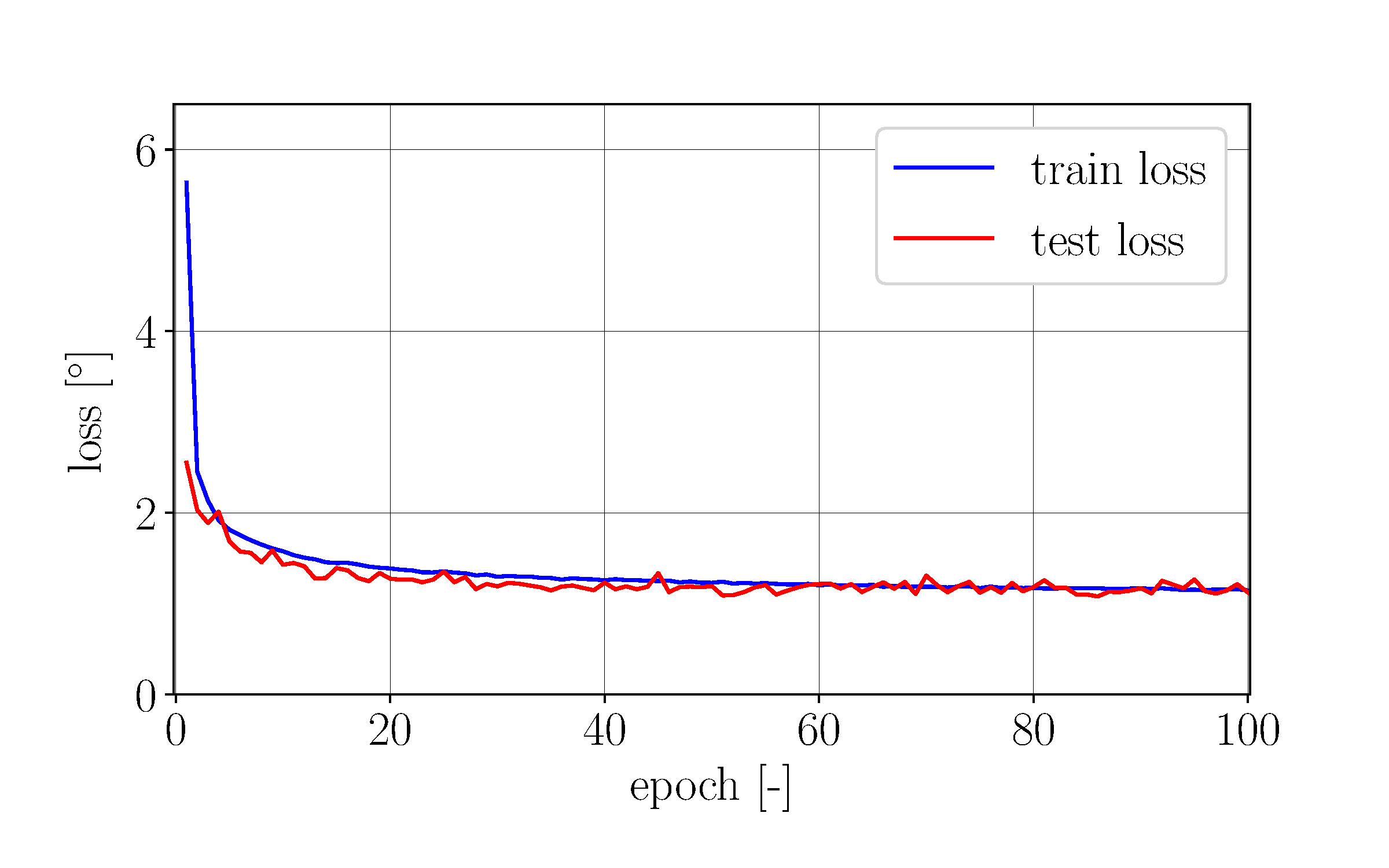}
	\caption{The figure shows the training and test loss over 100 epochs. The gap between train and test loss is small and no overfitting is apparent from the loss curves. The final training and test loss achieved, corresponds to a combined root-mean-square error in $\alpha$ and $\beta$ of $\unit[1.15]{^\circ}$.}
	\label{fig:loss}
\end{figure}

The PyTorch framework \citep{pytorch} is used for model training. The AdamW optimizer is used to minimize the mean squared error. The model is trained for 100 epochs with a learning rate of 1e-3 and a batch size of 128. The data is shuffled before training and a GPU (Nvidia Titan X Pascal) is used to execute the training. The evolution of the train and test loss over all epochs is shown in Figure \ref{fig:loss}.
Variations of the parameters defining the model architecture, namely the number of channels in each convolution, the convolution and pooling kernel sizes and the number of linear units were also considered, with no significant improvement in prediction accuracy for a network of similar size.

\vspace{6pt}
\subsection{Control}\label{sec:Control}
The control approach for the spherical robotic arm is introduced in this section. The focus of this work lies on the sensing method and therefore only a simple control strategy is employed. A more elaborate, i.e. model-based approach is documented in \cite{RoboSoft2021}. 

The control approach relies on a cascaded control architecture, similar to the one presented in \cite{HOFER2020102369}, with an outer control loop for the slower motion dynamics of the robotic arm and three independent, inner control loops for the faster pressure dynamics. Based on the sensory feedback of $\alpha$ and $\beta$, the position controller computes the pressure setpoints, which are the inputs to the inner control loops. The sensor feedback is either provided by the motion capture system or by the vision-based sensing approach presented in Section \ref{sec:CameraBasedSensing}. The control inputs required to track the setpoints $\alphaSP$ and $\betaSP$ are computed based on two decoupled proportional-integral controllers,
\begin{linenomath}
\begin{align}
	u_{\alpha}&=K_{\text{P}}(\alphaSP-\alpha) + K_{\text{I}}\int (\alphaSP-\alpha) \ dt\\
	u_{\beta} &= K_{\text{P}}(\betaSP-\beta) + K_{\text{I}}\int (\betaSP-\beta) \ dt,
\end{align}
\end{linenomath}
with $K_{\text{P}}$ and $K_{\text{I}}$ denoting the proportional and integral gains. The two control inputs are mapped to the three actuator pressure setpoints by applying the following two-part procedure. First, the control inputs $u_{\alpha}$, $u_{\beta}$ are aligned with the actuators by applying the following linear transformation, which is based on the actuator geometry (compare Figure \ref{fig:angleConvention}),
\begin{linenomath}
\begin{align}
	\pABSP&=-\frac{1}{\sqrt{3}}u_{\alpha} - u_{\beta}\\
	\pBCSP&=\frac{2}{\sqrt{3}}u_{\alpha},
\end{align}
\end{linenomath}
where $\pABSP$ corresponds to the pressure setpoint difference between actuators A and B and $\pBCSP$ to the difference between actuators B and C, respectively. Secondly, the relative pressure setpoint differences between two actuators are allocated to the absolute pressure setpoints by the following set of equations originating from \cite{RoboSoft2021},
\begin{linenomath}
\begin{align}
	\pASP&=\text{max} \{\pBar, \pBar+\pABSP, \pBar+\pABSP+\pBCSP\}\\
	\pBSP&=\text{max} \{\pBar, \pBar+\pBCSP, \pBar-\pABSP\}\\
	\pCSP&=\text{max} \{\pBar, \pBar-\pBCSP, \pBar-\pABSP-\pBCSP\}.
\end{align}
\end{linenomath}
Thereby, $\pBar$ is defined as a lower pressure level of all three actuators,
\begin{linenomath}
\begin{equation}
	\pBar = \text{min}\{\pASP,\pBSP,\pCSP\}.
\end{equation}
\end{linenomath}
The validity of the second step can be verified by computing the pressure differences between actuator A and B and similarly for B and C and performing the required case distinctions. The lower pressure level $\pBar$ can be interpreted as a mean to adjust the unidirectional stiffness of the robotic arm (see \citep{HOFER2020102369} for more details). 

The pressure setpoints for actuators A, B and C are tracked by three (independent) proportional-integral controllers,
\begin{linenomath}
\begin{align}
	\uA&=\tilde{K}_{\text{P}}(\pASP-\pA) + \tilde{K}_{\text{I}}\int (\pASP-\pA) \ dt\\
	\uB&=\tilde{K}_{\text{P}}(\pBSP-\pB) + \tilde{K}_{\text{I}}\int (\pBSP-\pB) \ dt\\
	\uC&=\tilde{K}_{\text{P}}(\pCSP-\pC) + \tilde{K}_{\text{I}}\int (\pCSP-\pC) \ dt,
\end{align}
\end{linenomath}
with $\tilde{K}_{\text{P}}$ and $\tilde{K}_{\text{I}}$ denoting the proportional and integral gains of the pressure controllers. The three controllers are executed on an embedded hardware at a higher rate than the position controller.

\section{Results}\label{sec:Results}

The results of the experimental evaluation of the method proposed are presented in this section. The results of the real-time prediction of the two angles are presented in the first part. The closed-loop experiments relying on the feedback from the camera-based sensing approach are presented in the second part. 

The ONNX Runtime framework\footnote{\url{https://onnxruntime.ai}} is used to reduce inference time of the neural network and provide a prediction of $\alpha$ and $\beta$ at a rate of $\unit[30]{Hz}$. The multithreaded software application, including model inference, is executed on a standard laptop computer (Intel Core i7 CPU, 2.8GHz). The position controller is executed at $\unit[50]{Hz}$, where the previous prediction of the angles is used for intermediate executions of the controller. The pressure controllers are executed at $\unit[1000]{Hz}$ on an embedded platform (STM32 Nucleo-144 development board with STM32F413ZH MCU from STMicroelectronics). All actuator pressures are measured by pressure sensors (8230 from Burkert) and the outputs of the pressure controllers are applied to proportional valves (MPYE-5-1/8-HF-010-B from Festo) adjusting the air pressures. The lower pressure level is set to $\bar{p}=\unit[1.02]{bar}$. Communication between the main application and the embedded platform is realized by means of serial communication. The interested reader is referred to the video attachment to gain an impression of the experiments conducted (\url{https://youtu.be/CldCKhukqqQ}).



\vspace{6pt}
\subsection{Real-Time Prediction}
\begin{figure}
\center
\includegraphics[width=16cm]{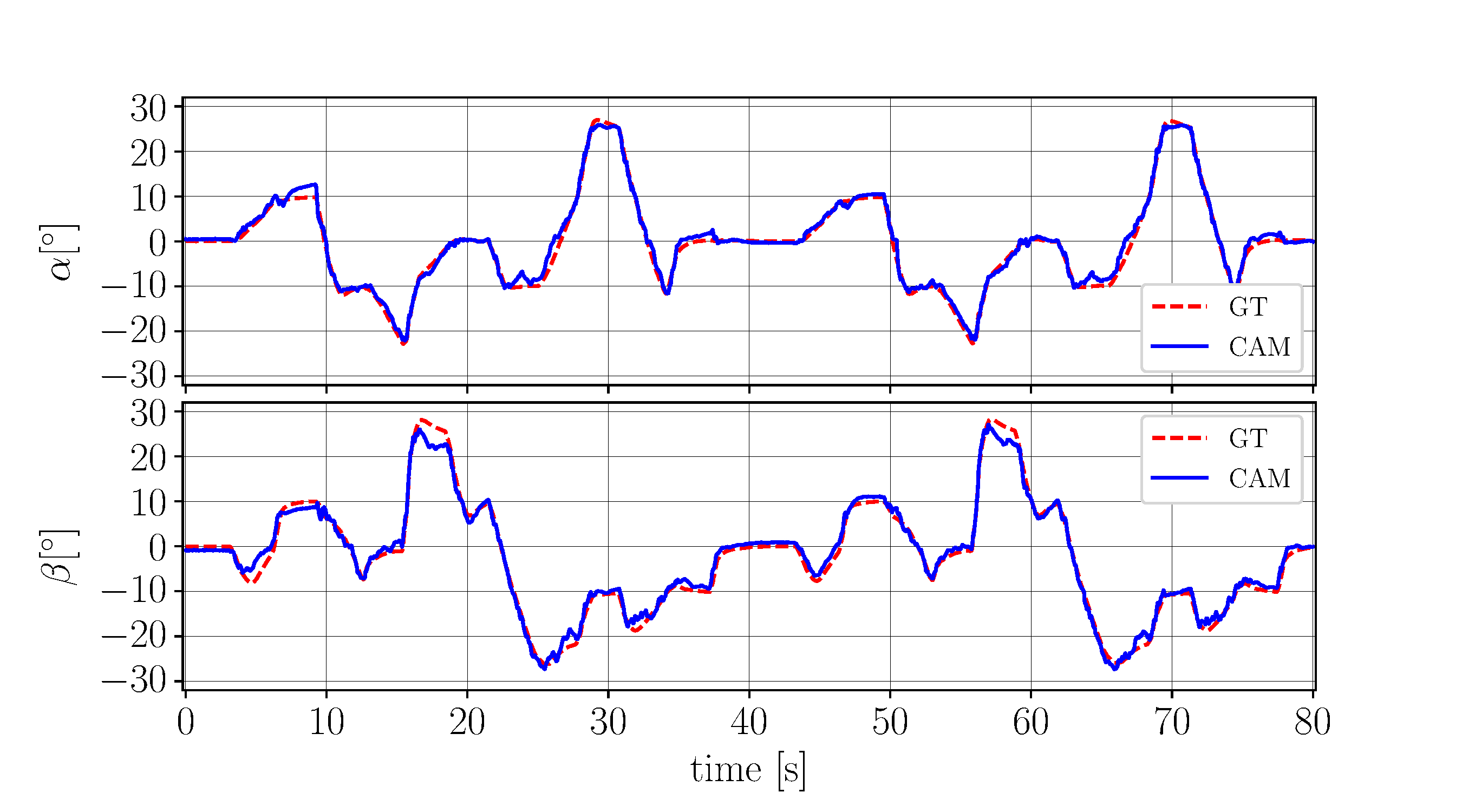}
\caption{The figure shows the rotational degrees of freedom $\alpha$ (top plot) and $\beta$ (bottom plot) over time. The camera-based prediction (CAM) is shown in blue (solid line) and the ground truth (GT) from the motion capture is depicted in red (dashed line). No filtering is applied to the output from the network. Slight deviations can be seen for large values of $\beta$. The deviations between the two repetitions of the trajectory are comparable.}
\label{fig:realTimePrediction}
\end{figure}

The robotic arm is commanded to track a series of steps and ramps in the $\alpha$ and $\beta$-directions, relying on feedback from the motion capture system. The range of frequencies considered corresponds to the range of frequencies in which the system typically operates. The trajectory is repeated once to investigate repeatability. The camera-based sensing approach is executed in real-time and the results are shown in Figure \ref{fig:realTimePrediction} showing the network prediction along with ground truth.

The root-mean-square error between prediction and ground truth is $\unit[1.1]{^\circ}$ in the $\alpha$ and $\unit[1.4]{^\circ}$ in the $\beta$-direction. The repetitive nature of the deviations indicates that there is a systematic trend which the network employed fails to capture. However, the similarity between the two realizations implies that a hardware limitation is not the cause of the  deviations in the predictions. The variations of the parameters defining the network architecture mentioned in Section \ref{sec:modelLearning} resulted in similar performance with deviations occurring in different regions of the $\alpha$-$\beta$-plane. This indicates that the network size is currently the limiting factor and improvements in one region are only possible at the cost of an accuracy degradation in another region.
\vspace{6pt}
\subsection{Vision-based Control}
\begin{figure}
\center
\includegraphics[width=16cm]{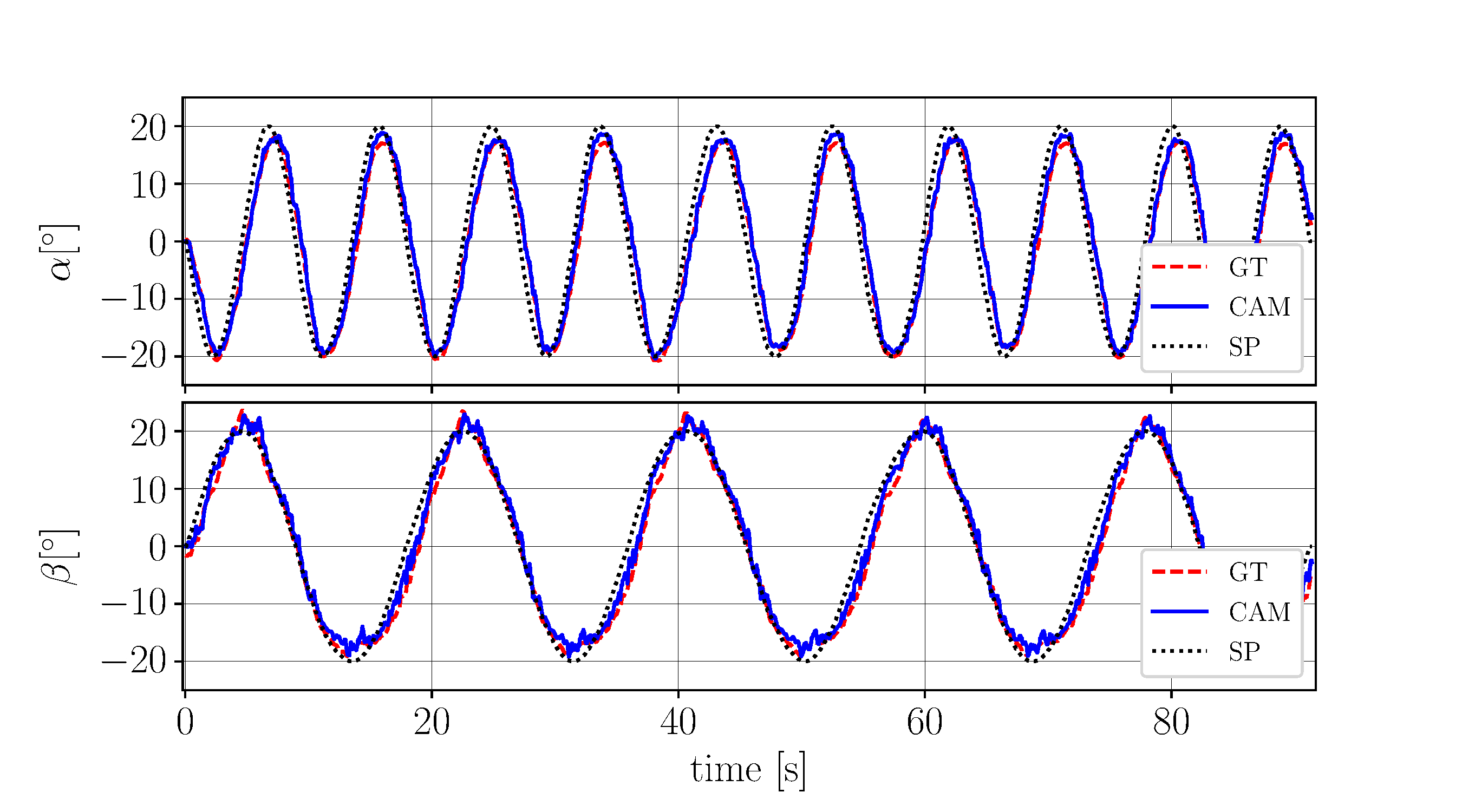}
\caption{The figure shows the results of a tracking experiment with the camera-based sensing approach used as sensory feedback. The angle $\alpha$ is shown in the top plot and $\beta$ is shown in the bottom plot. The camera-based predictions are shown in blue (solid line) and the ground truth from the motion capture in red (dashed line). The commanded setpoint is shown in black (dotted line).}
\label{fig:realTimeControl}
\end{figure}

The results of the closed-loop control experiment relying on feedback from the vision-based sensing approach are discussed in this section. The robotic arm is commanded to track sinusoidal setpoint trajectories in $\alpha$ and $\beta$-directions. Ground truth provided by the motion capture system is solely used for evaluation purposes. The results shown in Figure \ref{fig:realTimeControl} demonstrate the feasibility of the sensing approach for closed-loop control of the robotic arm. The root-mean-square error between prediction and ground truth is $\unit[1.1]{^\circ}$ in the $\alpha$ and $\unit[1.5]{^\circ}$ in the $\beta$-direction. Slight oscillations can be observed for large absolute values of $\beta$, which contribute to the larger error in the $\beta$-direction.

\section{Discussion}\label{sec:Discussion}

This paper presents a vision-based sensing approach for a soft robotic arm made from fabric. The camera integration into inflatable bellow actuators has been discussed with three actuators being used to control a spherical robotic arm. An end-to-end deep learning approach relying on a shallow convolutional network is employed and trained with ground truth data from a motion capture system to map the camera images to the two rotational degrees of freedom of the robotic arm. The resulting method is computationally lightweight and can be deployed in real-time on a standard laptop computer providing predictions of the two angles at rate of $\unit[30]{Hz}$ with an accuracy of $\unit[1.25]{^\circ}$. The reliability of the vision-based sensing approach has been demonstrated by closed-loop control experiments relying on the sensory feedback from the camera-based sensing approach.

There are several possible extensions to the approach presented: The sensing pipeline is identified for a fixed lower pressure level $\bar{p}$, which is related to joint stiffness. Therefore, the lower pressure level is likely to affect the actuator deformation and hence the images observed by the internal cameras. An extension of the current work would be to feed $\bar{p}$ as an input to the network and train it for different values of $\bar{p}$. Secondly, the robotic arm is commanded to track positions during the data collection and evaluation experiments. A subject for future work is an investigation of the method presented for deployment during interactive applications with disturbances acting on the movable link. Thirdly, the information stream provided by the cameras is currently only used for sensing the rotational degrees of freedom. An interesting extension would be to use the camera images to identify aging phenomena of the bellow actuators or to detect damages in the actuators, both changing the actuator deformation observed by the cameras. Finally, the sensing pipeline presented relies on three cameras (one in each actuator) to predict the rotational degrees of freedom. However, a change in both angles, in the $\alpha$ and $\beta$-directions causes a variation of each of the camera images. Hence, it is likely that the orientation of the robotic arm would be observable with only one or two cameras. This would simplify the hardware setup, but might come at the cost of reduced prediction accuracy.

In order to further improve the prediction accuracy, larger network architectures might be required. The repeatability of the current deviations indicates that the physical limitation of the sensing approach is not yet reached and better prediction accuracy is possible at the cost of larger networks and correspondingly higher computational costs for training and inference. Furthermore, we only investigated algebraic mappings from camera images to output angles without any previous state dependency. The use of e.g. recurrent neural network architectures to rely on past predictions to capture time dependent effects in the actuator deformation might be another means to improve the performance of the sensing approach presented.

\section*{Author Contributions}

Conceptualization, MH, CS and RD; Methodology, MH, CS and RD; Software, MH, CS; Validation, MH; Formal Analysis, MH, CS; Investigation, MH, CS; Resources, MH, CS and RD; Data Curation, MH; Writing - Original Draft Preparation, MH; Writing - Review and Editing, MH, CS and RD; Visualization, MH; Supervision, RD; Project Administration, MH, CS and RD; and Funding Acquisition, MH, CS and RD. All authors contributed to the article and approved the submitted version.

%
%
\section*{Acknowledgments}

The authors would like to thank Matthias Müller for his support with the camera setup and Jasan Zughaibi for his contribution to the hardware setup and sharing his knowledge on actuator behavior. Further thanks go to Julian Zilly for providing the GPU infrastructure to train the models and Michael Egli and Daniel Wagner for their support with the hardware.

%
%
\bibliographystyle{frontiersinSCNS_ENG_HUMS} 
\bibliography{bibliography}

\end{document}